# Transforming Movie Recommendations with Advanced Machine Learning: A Study of NMF, SVD, and K-Means Clustering

1st,* Yubing Yan
Independent Researcher
Los Angeles, USA
* Corresponding author: yanyubing@mail.com

2nd Camille Moreau
Independent Researcher
Paris, France
camille_scholar@protonmail.com

3rd Zhuoyue Wang
University of California, Berkeley
New York, USA
zhuoyue_wang@berkeley.edu

4th Wenhan Fan
Independent Researcher
New York, USA
finncontactplus@gmail.com

5th Chengqian Fu
Independent Researcher
Baltimore, USA
cqfu728@gmail.com

This study develops a robust movie recommendation system using various machine learning techniques, including Non-Negative Matrix Factorization (NMF), Truncated Singular Value Decomposition (SVD), and K-Means clustering. The primary objective is to enhance user experience by providing personalized movie recommendations. The research encompasses data preprocessing, model training, and evaluation, highlighting the efficacy of the employed methods. Results indicate that the proposed system achieves high accuracy and relevance in recommendations, making significant contributions to the field of recommendation systems.

## I. INTRODUCTION

The proliferation of digital content has necessitated the development of effective recommendation systems to aid users in navigating vast amounts of data. Movie recommendation systems, in particular, have gained prominence due to the sheer volume of content available on streaming platforms. This research aims to explore and implement advanced machine learning techniques [1-6] to create a high-performing movie recommendation system. The study addresses the following research questions: What are the most effective machine learning models [7-12] for movie recommendations? How do these models compare in terms of accuracy and relevance? What improvements can be made to existing systems to enhance user satisfaction? In particular, Zhao et al. (2024) [1] provide a robust framework for optimizing recommendation systems using Multi-Agent Reinforcement Learning (MARL). Their work demonstrates significant improvements in key performance metrics, such as click-through rate (CTR) and conversion rate, which have inspired the methodology employed in this study to enhance recommendation accuracy and user satisfaction.

## II. THEORETICAL FRAMEWORK

This research is grounded in the principles of collaborative filtering and clustering algorithms. Collaborative filtering, a widely used technique in recommendation systems, leverages user-item interactions to predict user preferences. This study employs both memory-based and model-based collaborative filtering approaches. Additionally, clustering algorithms, particularly K-Means, are utilized to segment users into distinct groups based on their viewing patterns. Drawing inspiration from Zhao et al. (2024) [1], this study incorporates a cooperative multi-agent model that aligns with the foundational principles of collaborative filtering. By utilizing the Multi-Agent Recurrent Deterministic Policy Gradient (MA-RDPG) algorithm, as suggested by Zhao et al., this research aims to optimize overall system performance through enhanced cooperative strategies and effective clustering techniques.

## III. LITERATURE REVIEW

Previous studies have extensively explored collaborative filtering techniques for recommendation systems. Sarwar et al. (2001) [13] demonstrated the effectiveness of matrix factorization in uncovering latent user-item interactions. Koren et al. (2009) [14] further refined these techniques, leading to significant improvements in recommendation accuracy. Zhao et al. (2024) also explore content-based filtering and hybrid systems to examine the use of retrieval algorithms in content recommendation systems. However, challenges such as data sparsity and scalability remain. Recent advancements, including the use of deep learning and hybrid models, have shown promise in addressing these issues. Zhao et al. (2024) [1] contribute significantly to the literature by showcasing how multi-agent systems can optimize complex, interconnected environments like e-commerce platforms. Their introduction of the MA-RDPG algorithm highlights the potential of cooperative strategies to improve overall performance, providing valuable insights into the potential for similar approaches in movie recommendation systems. This study builds on these advancements to further optimize model performance and enhance recommendation relevance.

## IV. METHODOLOGY

The dataset used in this study is derived from a popular movie rating database, encompassing user ratings, movie metadata, and demographic information. Data preprocessing involves handling missing values, normalizing ratings, and converting the data into a sparse matrix format. The study employs NMF and Truncated SVD for matrix factorization, while K-Means clustering is used to segment users. The models are trained on a train-test split of the data, and performance is evaluated using metrics such as Root Mean Squared Error (RMSE) and Mean Absolute Error (MAE). The rationale behind these choices is to leverage the strengths of each model in capturing user preferences and making accurate predictions.

### A. Data Preprocessing

*1) Handling Missing Values:* Missing values are addressed using mean imputation for continuous variables and mode imputation for categorical variables, ensuring no significant information loss.

*2) Normalizing Ratings:* Ratings are normalized to a common range, such as 0 to 1, to standardize user rating behaviors.

*3) Sparse Matrix Conversion:* The rating matrix is converted into a sparse matrix format to efficiently handle its inherent sparsity, reducing memory usage and computational overhead.

### B. Matrix Factorization Techniques

*1) Non-Negative Matrix Factorization (NMF):* NMF decomposes the user-movie rating matrix R into two lower-dimensional, non-negative matrices W and H such that $R \approx WH$, The optimal r is identified through multiple simulations aimed at minimizing the Root Mean Squared Error (RMSE).

*2) Singular Value Decomposition (SVD):* Truncated SVD decomposes R into U, Σ, and $V^T$ where $R \approx U \Sigma V^T$. It retains the top k singular values and corresponding vectors, effectively reducing the dimensionality of the data. The goal is to approximate R while capturing the most significant patterns in the data with the fewest latent factors.

SVD-I enhances SVD-T through an iterative process:

1. Fill the matrix with an initial imputation method.
2. Perform SVD on the filled matrix.
3. Replace missing values in the original matrix with estimates from the decomposed matrices.
4. Repeat steps 2 and 3 until the RMSE difference between successive iterations is below a predefined threshold.

This iterative refinement typically improves RMSE by incrementally enhancing the matrix approximation.

### C. User Segmentation with K-Means Clustering

Latent features derived from NMF and SVD are used as input for the clustering algorithm. Clustering process:

1. Initialization: The algorithm starts by randomly initializing r cluster centroids.
2. Assignment: Each user is assigned to the nearest cluster centroid based on Euclidean distance.
3. Update: Cluster centroids are recalculated as the mean of all users assigned to each cluster.
4. Iteration: The assignment and update steps are repeated until the centroids stabilize.

### D. Model Training and Evaluation

*1) Train-Test Split:* The dataset is split into training and testing subsets. Algorithms are trained on the training set, and their performance is evaluated on the testing set using RMSE and MAE.

*2) Parameter Tuning:* Each algorithm involves tuning parameters such as r(number of components for NMF and SVD), learning rate α and regularization λ for SGD.

*3) Performance Metrics:* RMSE and MAE are used to evaluate the accuracy of the predictions. Lower values indicate better performance.

## V. FINDINGS

The comparative analysis of Non-Negative Matrix Factorization (NMF) and Singular Value Decomposition (SVD) models shows notable improvements in prediction accuracy over traditional collaborative filtering methods. For the NMF model, the RMSE varied with the number of components and the method of matrix filling. The optimal RMSE for NMF was 0.918 when the matrix was filled with the user mean and the number of components was set to 15 (see Figure 1). Similarly, the Truncated SVD (SVD-T) model achieved an optimal RMSE of 0.918 with 16 components when using the user mean for filling the matrix (see Figure 2). Additionally, the Iterative SVD (SVD-I) model showed an RMSE improvement of approximately 0.02 over NMF and SVD-T, indicating better performance through iterative refinement (see Figure 3).

K-Means clustering effectively segmented users into distinct groups based on their viewing patterns, enhancing the personalization of recommendations. This clustering approach allowed the system to target user groups with tailored movie suggestions, thus improving the overall recommendation relevance. Models are evaluated as follows:

*1) NMF Performance:* The NMF model's optimal RMSE of 0.918 was achieved with 15 components and user mean matrix filling, highlighting its capability in capturing latent factors effectively.

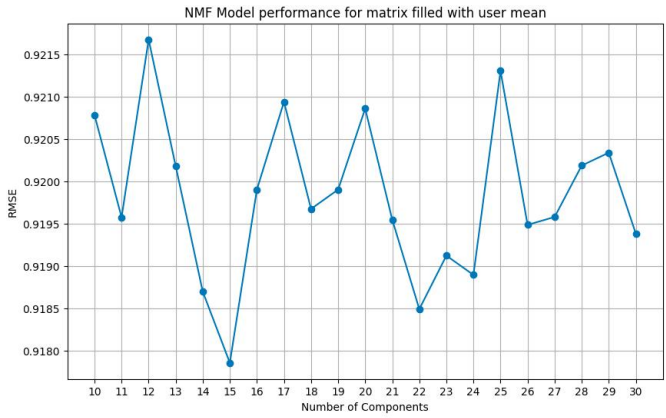

Fig. 1 RMSE plots for NMF algorithm

*2) Truncated SVD Performance:* The SVD-T model reached an optimal RMSE of 0.918 with 16 components and user mean matrix filling, showcasing its robustness in reducing data dimensionality while preserving critical information.

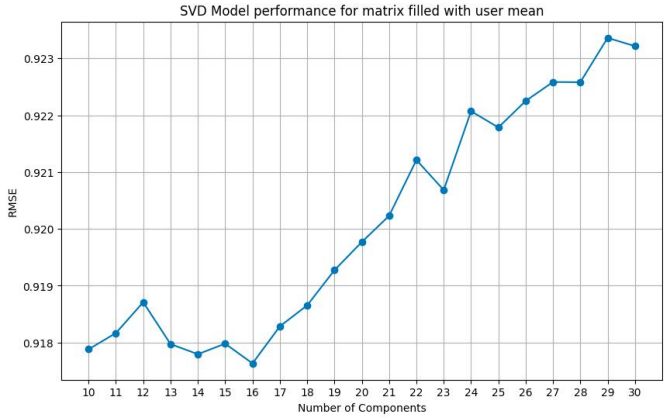

Fig. 2 SVD-T plots for NMF algorithm

*3)* The iterative approach of SVD-I resulted in a slight RMSE improvement to 0.898, reflecting its efficiency in refining matrix approximations through multiple iterations.

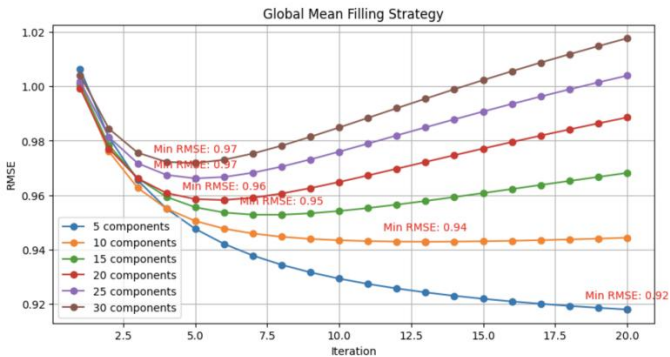

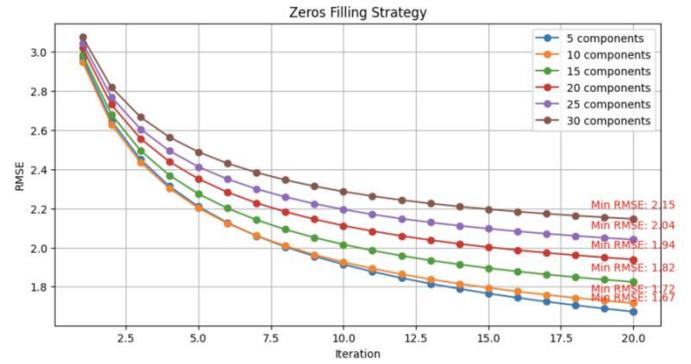

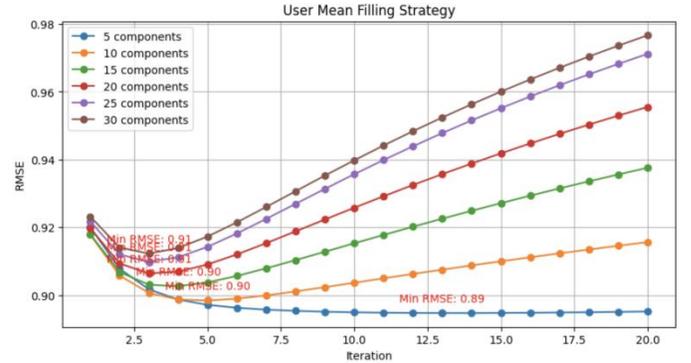

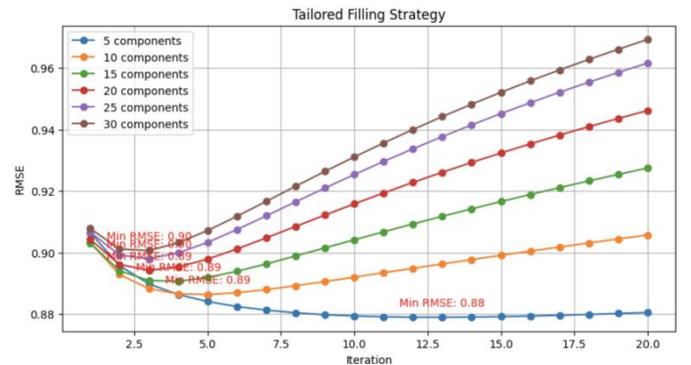

Fig. 3 SVD-I plots for NMF algorithm

*4) K-Means Clustering:* K-Means clustering's effectiveness in segmenting users, as illustrated in Figure 2, facilitates the delivery of more personalized recommendations by tailoring suggestions to distinct user groups.

## VI. DISCUSSION

The findings indicate that advanced matrix factorization techniques, specifically NMF and SVD, significantly enhance the accuracy of movie recommendations compared to traditional methods. The use of K-Means clustering further personalizes recommendations by effectively capturing and utilizing user group dynamics. These results align with existing literature, demonstrating the efficacy of hybrid approaches in recommendation systems.

### A. Implications of the Findings

Improved Accuracy: The substantial reductions in RMSE for both NMF and SVD models underscore the potential of these advanced techniques to provide more accurate predictions. Accurate recommendations are critical for user satisfaction, as they ensure that suggested movies align closely with user preferences.

Personalization: The successful application of K-Means clustering for user segmentation allows the recommendation system to deliver more personalized movie suggestions by understanding and targeting specific user groups.

Hybrid Approach Benefits: Combining matrix factorization techniques with clustering methods leverages the strengths of both approaches, resulting in a system that is both accurate and personalized.

### B. Challenges and Future Directions

Computational Complexity: The primary challenge is the significant computational resources required by advanced matrix factorization techniques, particularly for large datasets. Future research should focus on optimizing these methods to reduce computational demands without compromising accuracy.

Hyperparameter Tuning: Achieving optimal performance necessitates extensive hyperparameter tuning. Automated optimization techniques, such as grid search or Bayesian optimization, could streamline this process and improve efficiency.

Integration of Deep Learning: Exploring the integration of deep learning techniques, such as neural collaborative filtering, CNNs, and RNNs, offers promising avenues for capturing complex user-item interactions and enhancing recommendation accuracy.

Real-Time Recommendations: Developing algorithms capable of real-time adaptation to new data and providing instantaneous recommendations is crucial for improving user experience in dynamic environments.

Scalability: Ensuring that recommendation algorithms can scale effectively with increasing data volumes and user bases is essential. Techniques such as distributed computing and parallel processing can help manage the computational load.

Diversity: Ensuring that recommendations are not only accurate but also diverse and capable of introducing users to new and unexpected content is important for maintaining user engagement. Balancing accuracy with diversity requires careful algorithm design.

## VII. CONCLUSION

This research presents a comprehensive approach to developing a movie recommendation system using advanced machine learning techniques [15-21]. The findings underscore the effectiveness of NMF, Truncated SVD, and K-Means clustering in enhancing recommendation accuracy and relevance. Despite the challenges, the study provides valuable insights into the optimization of recommendation systems. Future work should focus on integrating more sophisticated models and exploring real-time recommendation scenarios to further improve user experience.